\documentclass{svproc}
\usepackage{url}

\usepackage{graphicx}
\usepackage{multicol}
\usepackage{footmisc}
\usepackage{tabularx}
\usepackage{booktabs}
\usepackage{cite}
\usepackage[separate-uncertainty=true]{siunitx}
\usepackage{float}
\usepackage{stfloats}
\usepackage{amsmath}
\usepackage{hyperref}
\usepackage{tikz}
\usepackage{eso-pic}
\usepackage[capitalize]{cleveref}
\crefname{equation}{Eq.}{Eqs.}
\crefformat{equation}{Eq.~(#2#1#3)}
\Crefformat{equation}{Equation~(#2#1#3)}

\begin{document}
\mainmatter              
\title{%
  \texorpdfstring
  {Walking on Roofs: Exploring the Potential of Walking Robots for Construction Work on Roofs}
  {Walking on Roofs: Exploring the Potential of Walking Robots for Construction Work on Roofs}
}
\titlerunning{Walking on Roofs}  
%
\author{Bjoern-Felix Dettmar \and Arne Roennau}
\authorrunning{B. Dettmar et al.} 
%
%
\institute{Machine Intelligence and Robotics Lab at Karlsruhe Institute of Technology (KIT), Germany\\
\email{dettmar@kit.edu}}

\maketitle 

\begin{tikzpicture}[remember picture,overlay]
  \node[
    anchor=north,
    yshift=-15mm,
    text width=\textwidth,
    align=center,
    font=\small\itshape
  ] at (current page.north)
  {Preprint of paper which appeared in the Proceedings of CLAWAR 2026,\\
29th International Conference on Climbing and Walking Robots and the 
Support Technologies for Mobile Machines,\\
Cranfield University, Cranfield, Beford, United Kingdome, 2026.};
\end{tikzpicture}

\begin{abstract} 
This paper investigates the feasibility of deploying quadruped walking robots for the automation of work in roof environments. While quadrupeds have demonstrated versatility across various domains, their large-scale deployment remains limited, partly due to lack of application-specific designs. Roof environments represent a novel and unexplored use case, combining high safety risks for human workers with repetitive, strenuous tasks that could benefit from robotic assistance. A dedicated test rig of a roof's surface was designed to evaluate the baseline performance of a commercial quadruped, the \emph{Unitree Go2}, in this new environment. Experiments revealed that standard ball feet are inherently inadequate for locomotion on sloped roofs: slippage increased quadratically with incline angle, get-up and lie-down sequences were only possible on small inclines, and critical failures already occurred regularly on moderate inclines of \SI{25}{\degree}. This work provides the first systematic assessment of quadruped locomotion in roof environments, highlighting both the potential and the current limitations of this application and establishes a baseline for future research. Effective solutions will require a combination of task-oriented foot designs, advanced contact mechanics and environment-specific control strategies, such as reinforcement learning for roof-adapted gaits.

\keywords{walking robots, quadruped, roof environment, construction, locomotion, friction, stability, motion capture}
\end{abstract}
\vspace{-1cm}
\begin{figure}[H]
    \centering
    \includegraphics[width=\textwidth]{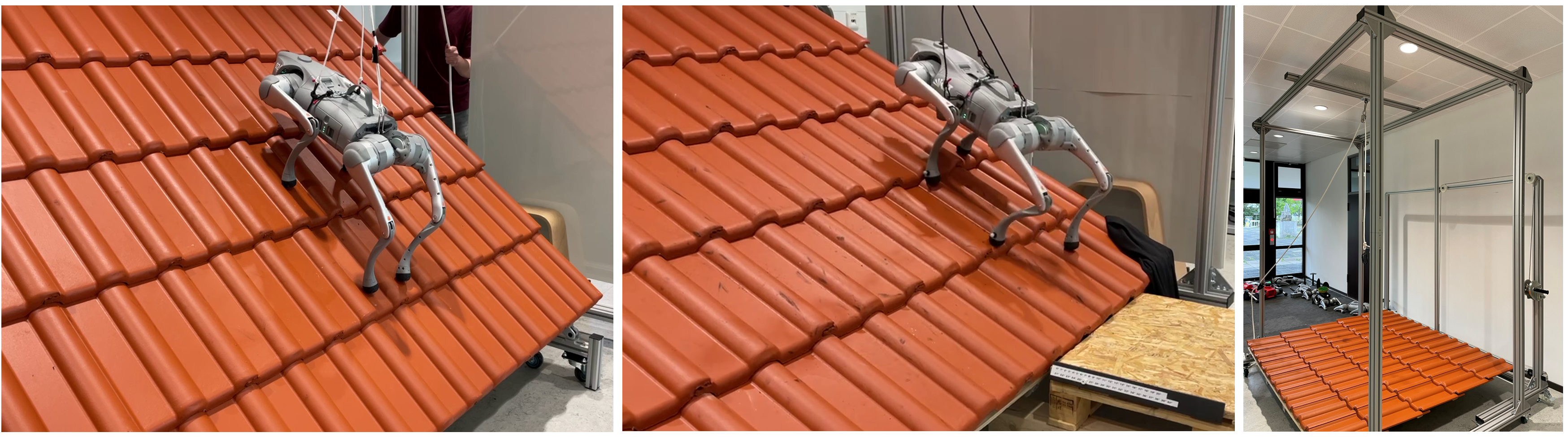}
    \caption{Examples of the \emph{Unitree Go2} on the test rig while balancing (left), walking (middle) and overview of the test rig setup (right).}
    \label{Figure_Intro}
\end{figure}

\section{Introduction}
The construction industry is among the least automated sectors with respect to robotics, despite its substantial economic relevance \cite{haghsheno_kunstliche_2024}. Within this domain, work on roofs represents a particularly hazardous area. Tasks such as construction, maintenance, installation, and inspection expose workers to significant risks, with slipping and falling accidents constituting a major cause of fatalities \cite{health_and_safety_executive_health_2020}. At the same time, many tasks on roofs are physically demanding and repetitive, providing a strong motivation to explore robotic assistance. In Germany alone, the roofing industry employed over 99,000 workers across more than 13,500 companies in 2023, generating a sales volume of nearly 12.9 billion euros \cite{zentralverband_des_deutschen_dachdeckerhandwerks_geschaftsbericht_2023}. This highlights both the scale of the sector and the potential impact of automation. 

Quadruped robots have become an increasingly active research field over the past decade and have consequently attracted growing interest from the industrial sector \cite{majithia_design_2024}. For commercial applications, both academia and industry have predominantly proposed sectors such as \emph{Inspection \& Maintenance}, \emph{Search \& Rescue}, and \emph{Military \& Surveillance} \cite{li_quadruped_2025}. Although these tasks have been proposed for a long time and quadruped capabilities have rapidly advanced, commercialization and deployment in these applications remain limited. This is exemplified by \emph{Boston Dynamics}, one of the earliest and best-known developers of quadrupedal robots, which reported deploying only about 2,100 units of their flagship model \emph{Spot} in industrial settings by 2023 \cite{boston_dynamics_retrospective_nodate}. This is in strong contrast to the rapid development of quadrupedal capabilities. Today, quadrupeds exhibit high mobility, adaptability in complex environments, and increasing energy efficiency \cite{li_quadruped_2025, taheri_study_2023}. Furthermore, with the emergence of deep learning, quadrupeds are becoming progressively more autonomous: their gaits are increasingly adaptable, and they can perform intelligent real-time monitoring and decision-making based on sensor data \cite{wang_learning_2024}. Thus, to enable quadruped deployment at scale, novel fields of applications with well defined tasks and challenges should be investigated. 

\begin{table}[h]
\centering
\caption{Potential applications for quadruped robots in roof environments.}
\begin{tabularx}{0.8\textwidth}{p{4cm} X}
\toprule
\textbf{Category} & \textbf{Tasks} \\
\midrule
Manipulation & 
Refurbishment and re-roofing \cite{health_and_safety_executive_health_2020} \newline
Maintenance and cleaning \cite{health_and_safety_executive_health_2020} \newline
Painting and protection \cite{highspecroofing_8_2024} \newline
Stripping and dismantling of roofs \cite{health_and_safety_executive_health_2020} \newline
Cutting overhanging vegetation \cite{highspecroofing_8_2024} \newline
Snow removal \cite{highspecroofing_8_2024}
\\
Precision & 
Installation of ancillary equipment \cite{health_and_safety_executive_health_2020} \newline
Chimney sweep
\\
Transport & 
Roof materials \cite{health_and_safety_executive_health_2020} \newline
Plant watering
\\
Information & 
Inspection and surveying \cite{health_and_safety_executive_health_2020}
\\
\bottomrule
\end{tabularx}
\label{Table_Roof_Task}
\end{table}

A potential new environment for quadruped robots to operate in is presented by roofs, offering a unique set of challenges and tasks. Quadrupeds could be deployed in roof environments for a multitude of applications, which are presented in \cref{Table_Roof_Task}. Residential buildings often feature moderate to steep inclines (\SIrange{25}{60}{\degree}), smooth roof tile surfaces, and regularly repeating structural elements such as convex and flat areas or tile row steps, making locomotion inherently difficult \cite{blas_dachformen_nodate}. Conversely, the repetitive nature of standardized roof tiles provides an opportunity to tailor quadrupeds feet specifically for this environment. 

\noindent
This work provides:
\begin{itemize}
    \item The first systematic experimental evaluation of quadruped locomotion in tiled roof environments.
    \item Quantitative failure thresholds for balanced standing, get-up/lie-down motions, and incline walking establishing a baseline for future research.
    \item A motion-capture-based methodology for objectively observing foot slippage during inclined locomotion.
\end{itemize}
\section{Evaluating Quadruped Locomotion in Roof Environments}
In roof environments, quadruped locomotion performance is primarily determined by two factors: friction and stability. To objectively evaluate and compare quadruped locomotion, an accurate measurement methodology is required. In this work the main focus is placed on friction as the primary performance indicator, while loss of stability, referred to as critical failure, marks the upper performance limit when the robot falls. Friction can be estimated by measuring slippage on a defined incline angle. This necessitates an external measurement system. For this purpose, the \emph{OptiTrack} system -- a motion capture system based on infrared light and reflective markers -- is employed. The system can be calibrated to achieve a measurement accuracy of below \SI{1.0}{\milli\meter} \cite{optitrack_prime_nodate}.

The feet of a quadruped robot are the primary, and often only, components that directly interact with the environment \cite{li_foot_2023}. While other end-effectors in robotics, such as hands and grippers, have received considerable attention, the development of non-humanoid feet has received less focus \cite{catalano_adaptive_2022, eckert_adaptive_2016}. Currently, most major manufacturers and research institutes employ a simple round foot geometry \cite{ranjan_design_2024, kalogroulis_can_2024}. This design is widely adopted due to its simplicity, robustness, and reliable performance across diverse environments \cite{kalogroulis_can_2024}. Further, the ball-shaped design allows for a simplified contact modeling approach, reducing the simulated contact to a point or line \cite{eckert_adaptive_2016}. However, such feet provide a limited contact area, resulting in high stress concentrations and an increased risk of slippage, particularly on inclined surfaces, as rubber friction is area dependent \cite{ranjan_design_2024, kalogroulis_can_2024, richhariya_unravelling_2023}. Moreover, ball feet lack adaptability to varying terrains, restricting their effectiveness in complex environments \cite{ranjan_design_2024}. Due to these drawbacks and the highly difficult roof environment, the development of customized and environment adapted feet is worth investigating in future work. 
\section{Experimental Setup}
As a preliminary experiment the dynamic coefficient of friction $f$ of the ball feet was measured by using the internal force-torque sensor of a \emph{UR10e}. The foot is pushed onto the roof tile surface with a defined normal force of 60 N, which is approximately the average force applied to a single foot of the \emph{Go2} during walking. Next, the foot is dragged over the tile for a distance of \SI{100}{\milli \meter}, while the antagonistic friction force is measured. The experiment is repeated 10 times, both on flat and curved sections of the roof tile. The mean coefficient of friction is calculated through the ratio between measured tangential force $F_T$ and applied normal force $F_N$ depicted in \cref{Formula_Friction}. The dynamic coefficient was measured instead of the static coefficient, because due to the nature of the soft and hollow rubber feet of the \emph{Go2}, a static coefficient could not reliably be measured, as pre-slip tangential force were affected by progressive foot deformation.
\begin{align}
    \label{Formula_Friction}
    f = \frac{F_T}{F_N} = tan(\rho)
\end{align}
To objectively assess the performance of the \emph{Go2} (software version 1.4.4) and create a baseline for future comparison with custom feet, an adjustable test rig of a roof environment was developed (see \cref{Figure_Intro}). As surface material the \emph{Frankfurther Pfanne} was chosen, which is the most common type of roof tile in Germany \cite{bmi_deutschland_gmbh_produktdatenblatt-braas-frankfurter-pfanne_nodate}. With the \emph{Go2}, three core skills were examined: balancing on continuously increasing inclines, get-up/lie-down sequences, and incline walking. To create statistical significance every experiment is repeated ten times. For balancing, the robot was placed facing uphill on a \SI{0}{\degree} incline, which was gradually increased until balance was lost, revealing failure modes under static conditions. Get-up and lie-down performance was assessed on progressively steeper inclines in three orientations (uphill, downhill, sideways), with success defined as completing both actions in each configuration. Incline walking was evaluated by repeatedly ascending increasingly higher slopes while measuring slippage using the \emph{OptiTrack} system to track reflective markers positioned at the sides of the ball feet. Slippage of the front and rear feet, $S_{\mathrm{FLF/RLF}}$, was defined as the cumulative incremental marker displacement opposing the walking direction $X$. The incremental displacement was calculated as $\Delta x_i(t)=x_i(t+1)-x_i(t)$, and only backward movements were accumulated. To exclude movements during the swing phase, only increments with negligible vertical displacement, defined as $|\Delta z_i(t)|<z_{\mathrm{tol}} = 1.0\text{mm}$, where $\Delta z_i(t)=z_i(t+1)-z_i(t)$, were considered. The resulting slippage calculation is given in \cref{Formula_Slippage}. 
\begin{align}
\label{Formula_Slippage}
S_\text{FLF/RLF} &= \sum_{t=0}^{N-2} C_i(t) \cdot\,
\max\!\left[0,-\left(\Delta x_i(t)\right)\right], \hspace{0.3cm}
C_i(t)=
\begin{cases}
1, & |\Delta z(t)| < z_{\mathrm{tol}},\\
0, & \text{otherwise.}
\end{cases}
\end{align}
For high repeatability reasons, all trials were controlled via Python using the \emph{WebRTC} implementation provided by \emph{Severov} \cite{severov_github_nodate}. The robot was set to standard walking mode and a target speed of \SI{1.0}{\meter\per\second}. This speed was chosen, because it represents a realistic speed at which the robot should operate to be viable in real world applications.

\section{Baseline Results}
Overall, the results indicate that, without significant modifications, the \emph{Go2} is not suited for operation within roof environments. Manufacturer claims of slope-climbing capabilities of up to \SI{40}{\degree} were not achieved under the test conditions \cite{unitree_robotics_unitree_data_sheet_nodate}. 

The coefficient of friction for a rubber ball foot on a roof tile was measured at $1.144 \pm 0.1299$ on flat and $1.430 \pm 0.1298$ on curved sections of the roof. Using \cref{Formula_Friction}, this leads to respective maximum material specific incline angles $\rho$ of ($48.84 \pm 3.22$)\si{\degree} and ($55.03 \pm 2.44$)\si{\degree}.

For standing on an incline, the critical angle was measured at \SI{27.58 \pm 0.72}{\degree} on average. Failures were primarily caused by automatic limb repositioning, which appears to be triggered by slippage of individual limbs or atypical joint angles. 

Get-up and lie-down experiments revealed additional limitations. Full task success was achieved only at low inclines: up to \SI{11.0}{\degree} for head-up and head-down orientations, and up to \SI{7.0}{\degree} for the sideways orientation. In all cases, failures occurred when the center of mass (COM) exited the support polygon during the pre-programmed motion. 

\begin{figure}[b]
    \centering
    \includegraphics[width=\textwidth]{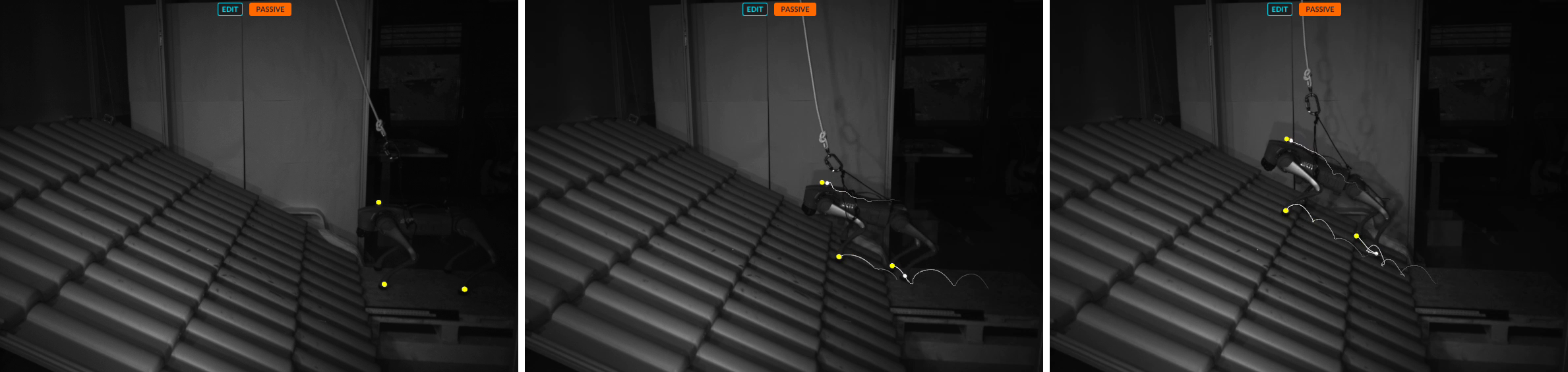}
    \caption{Motion capture view of the \emph{Unitree Go2} on the test rig at \SI{25}{\degree} incline angle.}
    \label{Figure_Videos_Walking_Optitrack}
\end{figure}

\begin{figure}[p]
    \centering
    \includegraphics[width=\textwidth]{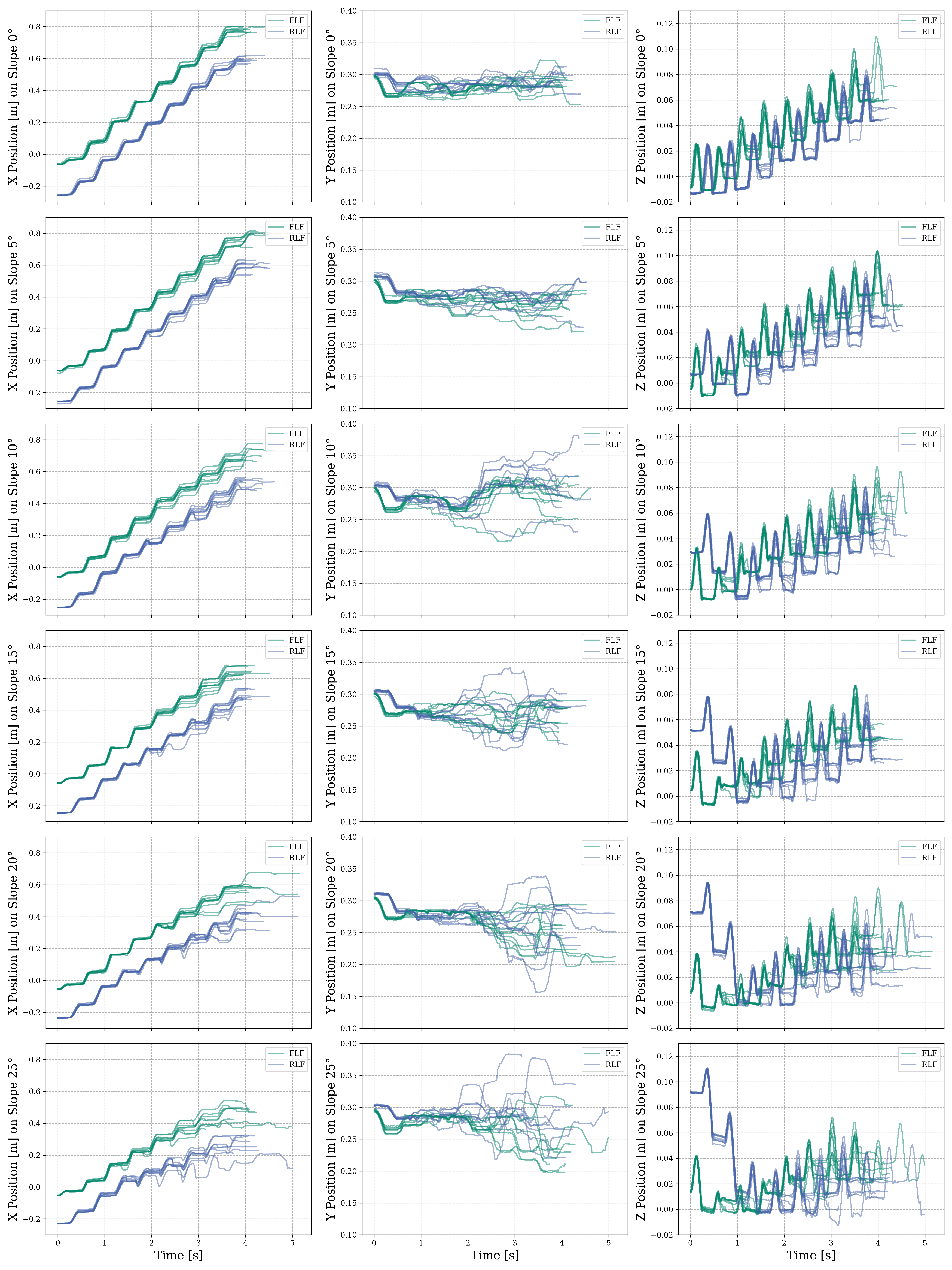}
    \caption{\emph{X}-, \emph{Y}- and \emph{Z}-trajectories of the \emph{Unitree Go2's} front left foot (FLF) and rear left foot (RLF) on incline angles \SI{0}{\degree} to \SI{25}{\degree}.}
    \label{Figure_Result_Walking_Trajectories}
\end{figure}

Incline walking performance was quantified by measuring mean slippage, calculated from positional data of reflective markers on front and rear feet. The robot was tested on inclines ranging from \SIrange{0}{25}{\degree}, with frequent falls observed at \SI{25}{\degree}. Experiment runs that ended in critical failure were excluded from later statistical analysis. Example trajectories of the experimental setup are provided in \cref{Figure_Videos_Walking_Optitrack}. The tracking data of each run in the roofs surface coordinate system is depicted in \cref{Figure_Result_Walking_Trajectories}.

The first column of \cref{Figure_Result_Walking_Trajectories} shows the absolute position of the front left foot (FLF) and rear left foot (RLF) in the walking direction (\emph{X}-direction). As the robot starts on flat ground, the first second of the plots is very similar, independent of the current inclination. The overall trajectory does change with higher inclines, though. On steeper slopes, slippage -- especially of the rear foot -- can be identified in areas where the absolute position decreases. The detected slippage is not a continuous sliding of the foot but rather an abrupt slip-off, often followed by an instantaneous follow-up step to regain ground contact. Furthermore, the plots show a greater variation in runtime. As the robot needs to rebalance more on higher inclinations, full standstill is sometimes reached later. The \emph{X}-direction plots also show that the roof environment makes the robot slip even on flat ground, due to the small steps between individual rows of tiles. When the robot steps on the edge of one of these steps, the foot can easily slip off, either during initial placement or at the start of the next step when the foot rotates forward and pushes off. Lastly, increasing the incline angle leads to a decrease in total traveled distance. This can be directly calculated from the diagram data, as the \emph{X}-direction is always set to be in-plane with the roof's surface, and distance measurements for calculation start only when all four feet are on the roof -- cutting off the part on the palette where movement is not directly in \emph{X}-direction for incline angles above \SI{0}{\degree}. 

The second column shows the deviation in \emph{Y}-direction, perpendicular to the \emph{X}-direction. These graphs indicate how well the robot maintains its given trajectory. Generally, the robot does not walk perfectly straight, even on flat ground. This can be seen in the beginning of the trajectory: starting on flat ground, both the FLF and RLF drift to the right, probably due to acceleration-induced imbalances. The roof environment further increases trajectory deviation, as the curved areas of the tiles promote slippage in \emph{Y}-direction. An increase in incline angle seems to correlate with greater trajectory deviation. Additionally, the robot might change its orientation on steeper inclines, as observed by the RLF drifting to the left side while the FLF drifts to the right side. 

The third column depicts the robot's gait pattern in \emph{Z}-direction. The horizontal areas of the graphs show phases in which the foot stays on the roof's surface. These horizontal areas increase with each first or second step, indicating that the foot is placed on the next row of tiles. Because the ground plane is always set to align with the lowest row of tiles, an increase in incline angle leads to an initial decrease in the \emph{Z}-position, as the \emph{Z}-axis points partially into negative \emph{X}-direction on flat ground. The increase in incline angle also correlates with a smaller increase in height, as the robot does not walk as far. Furthermore, the plots show greater variance, as the robot slips more often and deviates from its standard gait pattern. 

\begin{figure}[t]
    \centering
    \includegraphics[width=\textwidth]{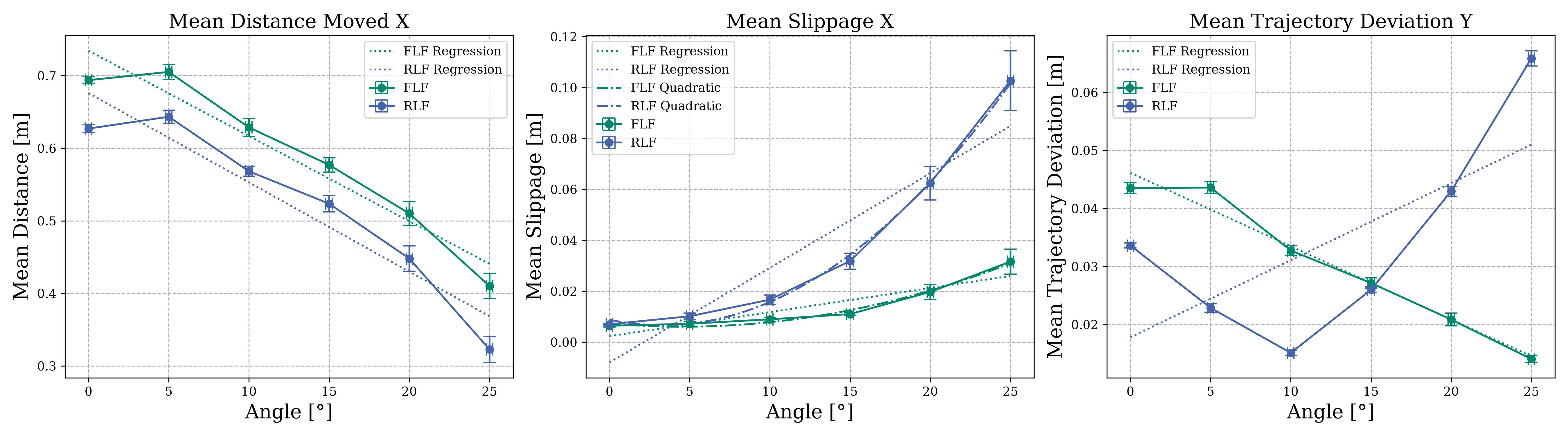}
    \caption{Mean distance traveled, slippage and trajectory deviation of the front left (FLF) and rear left foot (RLF), including linear regression models and a quadratic model for slippage.}
    \label{Figure_Walking}
\end{figure}

For better comparison the metrics mean distance moved in walking direction ($X$-direction), mean slippage in walking direction ($X$-direction) and mean trajectory deviation from expected trajectory ($Y$-direction) are derived from the tracking data and presented in \cref{Figure_Walking} for the inclination angles \SIlist{0;5;10;15;20;25}{\degree}. Regression analysis revealed a highly significant ($p < 0.0035$) linear relationship between incline angle and traveled distance, a highly significant ($p(F) < 0.001$) quadratic relationship between incline angle and slippage, but no explainable correlation between incline angle and trajectory deviation. A quadratic relationship for slippage is consistent with physical principles, as slippage corresponds to accelerated motion once static friction is overcome. Detailed information on regression results are presented in \cref{Table_Baseline_Regression} and a summary of all final results in \cref{Table_Final_Results_Overview}.
\begin{table}[t]
\centering
\caption{Linear regression analysis of mean distance moved in \emph{X}-direction, mean slippage in \emph{X}-direction and mean trajectory deviation in \emph{Y}-direction and quadratic regression of mean slippage in \emph{X}-direction.}
\begin{tabular}{lccccc}
\hline
\addlinespace[0.05cm]
\textbf{Variable} & \textbf{Linear Function} & \textbf{r} & \textbf{R\textsuperscript{2}} & \textbf{p} \\
\hline
\addlinespace[0.05cm]
mean\_dist\_FLF & $y = -0.01175x + 0.7343$     & -0.9680 & 0.9370 & 0.0015 \\
mean\_dist\_RLF & $y = -0.0123x + 0.6759$      & -0.9516 & 0.9056 & 0.0035 \\
mean\_slip\_FLF & $y = 0.0009464x + 0.002331$  & 0.9023  & 0.8141 & 0.0139 \\
mean\_slip\_RLF & $y = 0.003716x - 0.007981$   & 0.9286  & 0.8623 & 0.0075 \\
mean\_traj\_FLF & $y = -0.001264x + 0.04613$   & -0.9847 & 0.9696 & 0.0003 \\
mean\_traj\_RLF & $y = 0.001328x + 0.01781$    & 0.6860  & 0.4706 & 0.1324 \\
\hline
\addlinespace[0.05cm]
\textbf{Variable} & \textbf{Quadratic Function} & & \textbf{R\textsuperscript{2}} & \textbf{p(F)} \\
\hline
\addlinespace[0.05cm]
mean\_slip\_FLF & $y = 5.962 \cdot 10^{-5}x^2 - 0.0005x + 0.0073$ & &  0.986 & 0.00158 \\
mean\_slip\_RLF & $y = 0.0002x^2 - 0.0013x + 0.0088$ & & 0.997 & 0.000153 \\
\hline
\end{tabular}
\label{Table_Baseline_Regression}
\end{table}

\section{Discussion}
The purpose of this work was to evaluate the feasibility of using quadruped robots for automation of tasks in roof environments. For this, three core skills -- balancing, get-up/lie-down sequences and incline walking -- were evaluated on a test rig to create the first ever baseline for quadrupeds in this environment. The experiments provided an in-depth demonstration that quadruped robots like the \emph{Go2} are currently not suited for deployment within roof environments with frequent critical failures during incline walking at \SI{25}{\degree}. 

\begin{table}[t]
\centering
\caption{Comparison of all final experiment results.}
\begin{tabularx}{\textwidth}{p{5.5cm} X}
\toprule
\textbf{Parameter} & \textbf{Ball Foot} \\
\midrule
Friction coefficient flat & $1.144 \pm 0.1299$ \\
Friction coefficient curved & $1.430 \pm 0.1298$ \\
Standing on incline angle [\si{\degree}] & $27.58 \pm 0.72$ \\
Stand up/lie down (Head up) [\si{\degree}] & 11 \\
Stand up/lie down (Head down) [\si{\degree}] & 11 \\
Stand up/lie down (Sideways) [\si{\degree}] & 7 \\
Performance flat ground & Very high, practically no slippage \\
Performance \SI{0}{\degree} incline angle & Medium, slipping on curved area / tile steps \\
Max walking incline angle [\si{\degree}] & 20 \\
\bottomrule
\end{tabularx}
\label{Table_Final_Results_Overview}
\end{table}

Generally, the measured friction coefficients of the rubber ball feet suggest that theoretically sufficiently steep incline angles should be reachable. However, dynamic experiments reveal the opposite -- the robot does not come close to theoretic inclinations. There are multiple reasons that contribute to this gap. First on increasing inclines the COM shifts from front legs to rear legs, effectively reducing normal forces and thus friction at the front feet. This is supported by the standing on incline experiments, where failure was always preceded by slippage of one of the front limbs. Second, the robot controller does not change its behavior on incline. This means the robot tries to push off the ground as it would on flat terrain, contributing to higher tangential forces causing the resulting force vector to fall outside of the friction cone. This can lead to slipping even with sufficiently high friction coefficients. Third, as the rubber is relatively hard, dynamic impacts are not directly dampened, introducing stick-slip behaviors. Finally, controlled experiments with a constant normal force and small accelerations can not be fairly compared to the highly dynamic scenarios of robot locomotion, therefore, the calculated friction values should be viewed as theoretical maximum values that can only be approached under ideal conditions. The experiments highlight that roof locomotion is strongly limited by dynamic stability and control, but also by material friction.

Beyond the evaluation of a single platform, the experimental results allow several general design implications for quadruped locomotion in roof environments to be derived:
\begin{enumerate} 

\item \textbf{Dynamic locomotion performance cannot be predicted from friction measurements alone.}
The measured dynamic coefficients of friction suggest theoretically achievable incline angles above \SI{48}{\degree}, yet locomotion failed at substantially lower slopes. This gap highlights that dynamic locomotion is dominated by transient load redistribution, impacts, and controller behavior. Consequently, material friction tests alone are insufficient to evaluate locomotion suitability for steep environments.

\item \textbf{Center-of-mass redistribution emerges as a primary limiting factor on inclines.}
Failures during balancing were consistently preceded by slippage of the front limbs. As incline angle increases, the center of mass shifts toward the rear legs, reducing normal forces and therefore friction forces at the front feet. This indicates that incline locomotion requires active and incline-aware COM management strategies, rather than flat-terrain control policies.

\item \textbf{Discrete terrain features dominate slip events.} 
Slippage was frequently triggered by abrupt slip-off events at tile steps and curved sections, even on \SI{0}{\degree} incline. This suggests that roof locomotion is governed by discrete terrain interactions rather than purely planar slope mechanics. Foot placement strategies and terrain perception therefore become critical components for reliable locomotion.

\item \textbf{Standard quadruped control policies are not incline-aware.}
The evaluated controller applied similar push-off forces and gait timing across all inclines. This resulted in tangential forces exceeding the friction cone on steeper slopes. The findings indicate that environment-specific locomotion strategies are necessary, including slope-adaptive gait parameters and force modulation.

\item \textbf{Foot geometry is a critical hardware bottleneck.}
Ball-shaped feet exhibited systematic limitations due to their small contact area and lack of geometric adaptation to tiled surfaces. The results suggest that future roof-capable quadrupeds should employ feet with larger and adaptable contact areas, high-friction materials, and the ability to mechanically interlock with repetitive terrain structures.

\item \textbf{Get-up/lie-down motions must be terrain-aware.}
Get-up and lie-down motions failed at very low inclines because the pre-programmed sequences do not consider body orientation relative to gravity. Safe deployment therefore requires incline-aware recovery behaviors or dedicated level activation areas.

\end{enumerate}

Together, these findings indicate that successful deployment in roof environments will require a co-design approach combining terrain-aware control, perception-driven foothold selection, and environment-specific foot hardware.

Based on the identified design implications, multiple approaches should be investigated to increase quadruped performance within roof environments. The use of environment specific reinforcement learning is a promising first step to tackle some of the aforementioned issues, by teaching the robot to actively shift its COM to redistribute weight more evenly and reduce additional tangential forces to enforce actuation within the friction cone limits. Further, perception capabilities could be included to avoid problematic foot positions, like the steps between tiles, that promote slipping. The periodic nature of the tiled surface offers significant potential for locomotion, as strategic foothold selection could substantially reduce terrain difficulty. While performance can certainly be improved by environment specific training through methods like reinforcement learning, the quadratic correlation between slippage and incline angle will likely still make it necessary to increase friction through geometry or material changes. The point contacts limited adaptability and small contact area are inherently disadvantageous for locomotion in steep environments, as friction for rubber materials is surface area dependent \cite{richhariya_unravelling_2023}. Future robot feet for roof environments should be equipped with a larger, adaptable contact area, high friction materials, and actuated ankle joints to allow for more stable positioning of the body. 

When designing new feet, get-up and lie-down sequences also have to be kept in mind. Currently, the robot gets up by rolling over part of the ball foot. Depending on potential foot geometries, this might not be possible. Further, these sequences are not including information about body orientation and are thus prone to failure on incline. Either new, incline angle sensitive motions have to be trained or the robot is always activated on level platforms, like scaffolding. As failure during experiments already occurred on very low inclines, the latter option could prove safer. 

Despite the valuable insights gained, this study has several limitations. The \emph{Go2}, while commercially available is not a high end model and since its release, multiple high performance quadrupeds have been released. Further, all experiments were conducted using the standard walking gait of the \emph{Go2} at a single walking speed of \SI{1.0}{\meter \per \second}. Software updates have since integrated a new "AI walking mode" which in preliminary tests has proven to increase possible incline angles at least moderately. Lastly, all experiments have been conducted under ideal conditions, no additional debris, vegetation, weather or additional payloads were added. These aspects are likely to negatively impact quadruped performance and add uncertainty to surface properties. To allow for more general insights into the suitability of quadrupeds in roof environments, different robotic platforms should be tested under various conditions and with different settings.

Overall, roofs present a highly challenging new environment for quadrupeds, for which a combination of hardware and software adaptions will likely be necessary to enable automation of tasks.

\section{Conclusion}
Roof environments represent a compelling and largely unexplored application for robotics in an industry that has not yet undergone automation. Roof work combines high-risk conditions for humans with repetitive and physically demanding tasks that could benefit from robotic assistance. 

The experiments revealed that smooth, inclined, and structured roof surfaces pose significant challenges. Static tests demonstrated that an unmodified \emph{Unitree Go2} reaches critical failure before encountering the common incline angles of residential buildings, while dynamic tests showed a quadratic relationship between slippage and inclination. Further, measured material based coefficients of friction suggest that potentially much higher incline angles should be reachable, indicating control to be a strong limiting factor. Additionally, get-up and lie-down sequences proved unsafe on inclines. These results underscore that, without appropriate adaptations, quadrupeds are not yet capable of safe and reliable operation in roof environments. 

Future work should adopt an integrated approach that combines tailored locomotion strategies and perception with hardware adaptations to meet the specific demands. This includes the development of roof-specific gaits via reinforcement learning, supported by accurate modeling of friction and ground interactions, as well as modifications to the robots’ end-effectors by integrating foot actuation capabilities. The synergy between mechanical redesign and control adaptation significantly narrows the gap toward achieving autonomous quadrupedal navigation in roof environments.

%
%
\bibliographystyle{styles/bibtex/spmpsci_custom}
\bibliography{bib/references}
\end{document}